\documentclass[]{fairmeta}

\usepackage{graphicx}
\usepackage{booktabs}
\usepackage{amssymb}
\usepackage{orcidlink}

\usepackage{multirow}
\usepackage[table]{xcolor}
\usepackage{array}

\title{Efficient Universal Perception Encoder}

\author[1,*,\dagger]{Chenchen Zhu}
\author[1,*]{Saksham Suri}
\author[2,*]{Cijo Jose}
\author[2,*]{Maxime Oquab}
\author[2]{Marc Szafraniec}
\author[1]{Wei Wen}
\author[1]{Yunyang Xiong}
\author[2]{Patrick Labatut}
\author[2]{Piotr Bojanowski}
\author[1,\dagger]{Raghuraman Krishnamoorthi}
\author[1,\dagger]{Vikas Chandra}

\affiliation[1]{Meta Reality Labs}
\affiliation[2]{FAIR at Meta}

\contribution[*]{core contributor}
\contribution[\dagger]{project lead}

\abstract{
    Running AI models on smart edge devices can unlock versatile user experiences, but presents challenges due to limited compute and the need to handle multiple tasks simultaneously. This requires a vision encoder with small size but powerful and versatile representations. We present our method, \textbf{E}fficient \textbf{U}niversal \textbf{P}erception \textbf{E}ncoder (EUPE), which offers both inference efficiency and universally good representations for diverse downstream tasks. We achieve this by distilling from multiple domain-expert foundation vision encoders. Unlike previous agglomerative methods that directly scale down from multiple teachers to an efficient encoder, we demonstrate the importance of first scaling up to a large proxy teacher and then distilling from this single teacher. Experiments show that EUPE achieves on-par or better performance than individual domain experts of the same size on diverse task domains and also outperforms previous agglomerative encoders. We release the full family of EUPE models and the code to foster future research.
}

\correspondence{\email{chenchenz@meta.com}}

\metadata[Code]{\url{https://github.com/facebookresearch/eupe}}
\metadata[Model]{\url{https://huggingface.co/collections/facebook/eupe}}

\begin{document}

\maketitle
\section{Introduction}
\label{sec:intro}

Foundation vision encoders have made substantial progress in both architectures and training recipes. Popular architectures include convolutional neural networks \cite{he2016resnet, resnext, densenet, liu2022convnext} and vision transformers \cite{dosovitskiy2021vit, swin, deit}. They are trained either by full supervision \cite{kirillov2023sam, sam2, sam3}, weak supervision on text-image pairs \cite{radford2021clip, tschannen2025siglip, bolya2025perception}, or self-supervision \cite{oquab2024dinov2, mae, mocov3, beit}. They provide powerful feature representations for transfer to downstream vision tasks.
Meanwhile, downstream tasks are also evolving rapidly. Classical tasks include image understanding, such as image classification \cite{imagenet, objectnet, sun397} and image retrieval \cite{inaturalist, coco, flickr}, as well as dense prediction, e.g., segmentation \cite{ade, voc}, depth \cite{nyu, kitti}, and keypoint correspondence \cite{spair, navi}. Recently, vision-language modeling tasks are gaining popularity. Connecting a language model with a vision encoder has become a general paradigm for Visual Question Answering tasks. Cambrian-1 \cite{tong2024cambrian} groups these tasks into roughly four categories: OCR, vision-centric, knowledge, and general.

A single foundational vision encoder usually excels in one or two task domains. For example, encoders trained on text-image pairs such as CLIP \cite{radford2021clip}, SigLIP \cite{siglip, tschannen2025siglip}, and PEcore \cite{bolya2025perception} demonstrate strong performance in image understanding and vision-language modeling, yet their performance on dense prediction tasks often falls below expectations. DINO \cite{oquab2024dinov2, simeoni2025dinov3} and SAM \cite{kirillov2023sam} excel at dense prediction, but lack satisfactory vision-language capabilities. Consequently, downstream applications require the careful selection of a specific encoder to avoid performance degradation. Additionally, for use cases involving multiple domains, we either need to sacrifice computational efficiency to include multiple encoders or accept the performance tradeoff due to relying on a specific encoder.

\begin{figure}[t]
  \centering
  \includegraphics[width=0.65\linewidth]{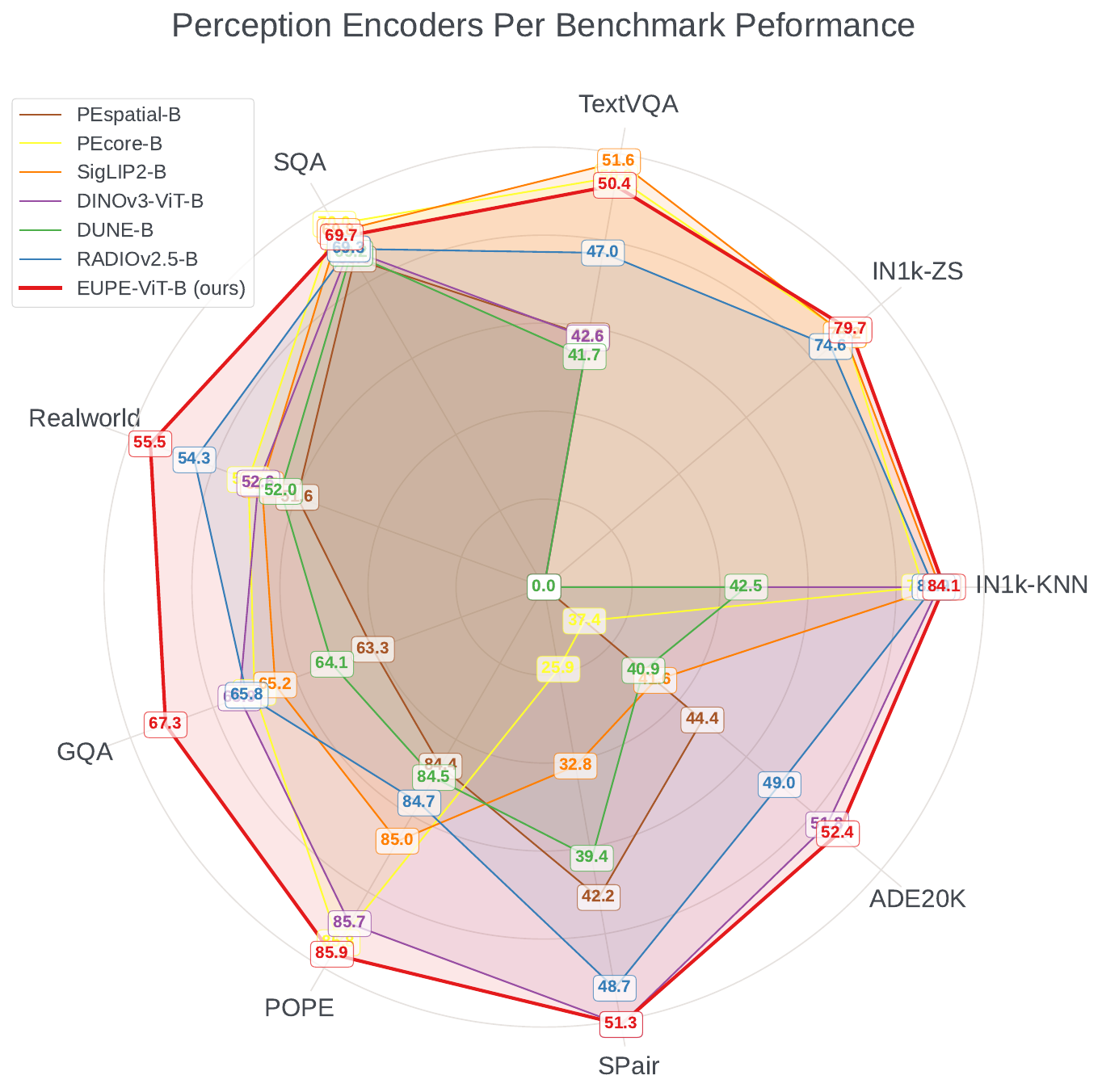}
  \includegraphics[width=0.19\linewidth]{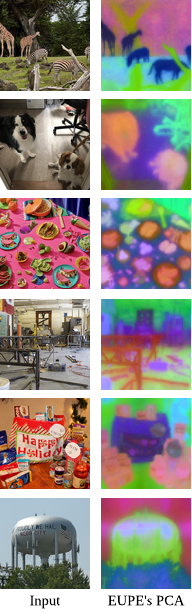}
  \caption{Applying our distillation recipe (EUPE) to ViT-B gives a well-balanced universal encoder that excels at diverse task domains compared to both ViT-B domain experts and existing agglomerative ViT-Bs. \textit{\textbf{Left}}: Performance on benchmarks across three task domains, higher the better. IN1k-ZS and IN1k-KNN are image understanding benchmarks on ImageNet1k. TextVQA, SQA, Realworld, GQA, POPE are vision-language modeling tasks. SPair and ADE20k are dense prediction tasks. We omit the IN1k-ZS score for models without text encoder (PEspatial-B, DINOv3-ViT-B, DUNE-B) and the IN1k-KNN score for models without class token output (PEspatial-B). \textit{\textbf{Right}}: Visualization of EUPE-ViT-B's feature by PCA projection into RGB space.}
  \label{fig:teaser}
\end{figure}

To address this issue, PE \cite{bolya2025perception} applies alignment tuning to intermediate layers, leading to three variations that excel at image understanding, dense prediction, and vision-language modeling, respectively. However, this still raises the question: \textit{can we agglomerate multiple domain capabilities into a single encoder?} RADIO \cite{ranzinger2024radio, heinrich2025radiov2} shows that this can be achieved through label-free knowledge distillation from multiple teacher models with the teachers being individual domain experts. Although it works well for large encoders (e.g., more than 300M parameters), we observe clear limitations when applying it to efficient backbones. As shown in Fig. \ref{fig:teaser} left, RADIOv2.5-B \cite{heinrich2025radiov2} has significant gaps compared to domain experts on dense prediction and VLM tasks.
On the other hand, efficient encoders are essential for personal super-intelligence on edge devices. Models running on them need to deal with limited compute resources and are often deployed in a multi-task setting. Therefore, developing a recipe for efficient universal encoders is fundamental to power versatile AI experiences for edge devices.

In this work, we study the pretraining recipe to produce efficient universal perception encoders. We discover that the principle to achieve universal capability on efficient encoders is first scaling up and then scaling down. Directly scaling down from multiple foundation teachers like in previous approaches cannot deliver satisfactory results because the efficient encoders do not have enough capacity to absorb various feature representations from foundation teachers into a universal representation directly. We propose the concept of proxy teacher which is a heavy model with enough capacity to unify the knowledge from multiple foundation teachers. This proxy teacher then transfers the learned universal knowledge to efficient students through distillation. To fully leverage the power of the proxy teacher, we distill the students from it with a longer fixed-resolution stage and shorter multi-resolution stage to accommodate the downstream tasks at various resolutions. Applying this recipe to efficient encoders leads to our \textbf{E}fficient \textbf{U}niversal \textbf{P}erception \textbf{E}ncoder (EUPE) family.

Experiments show that the proposed scaling-up and scaling-down distillation pipeline without additional bells and whistles can produce efficient universal encoders on-par or outperforming individual domain experts with the same size when zero-shot transferring to downstream tasks. For example, with the ViT-B architecture as shown in Fig. \ref{fig:teaser} left, our EUPE is on-par with image understanding experts like PEcore \cite{bolya2025perception}, SigLIP2 \cite{tschannen2025siglip}, and DINOv3 \cite{simeoni2025dinov3} on ImageNet-zeroshot and ImageNet-knn metrics, respectively. It is also on-par for even out-performing the dense prediction expert DINOv3 \cite{simeoni2025dinov3} on SPair \cite{spair} and ADE20k \cite{ade}. Compared to the vision-language modeling expert PEcore \cite{bolya2025perception} and SigLIP2 \cite{tschannen2025siglip}, it achieves significantly better performance on RealworldQA \cite{realworld}, GQA \cite{gqa} while maintaining at par performance on TextVQA \cite{textvqa}, SQA \cite{sqa}, and POPE \cite{pope}. Additionally, it outperforms existing agglomerative methods such as RADIO \cite{heinrich2025radiov2} and DUNE \cite{sariyildiz2025dune} by large margins on most benchmarks. Fig. \ref{fig:teaser} right visualizes EUPE-ViT-B's feature through PCA projection. Qualitatively, the feature can capture the semantic coherence (row 1\&2), fine granularity (row 3), complex spatial structure (row 4), and text awareness (row 5\&6) at the same time.

In summary, our main contributions include:
\begin{itemize}
  \item A simple scaling-up and scaling-down distillation recipe that produces powerful efficient universal perception encoders, outperforming existing agglomerative methods.
  \item A zoo of efficient model checkpoints with on-par or better performance than domain expert encoders on various downstream tasks for diverse on-device use cases under different computation budgets.
  \item A comprehensive study of the distillation recipe to share insights on training stages, teachers, and other hyperparameter choices.  
\end{itemize}
\section{Related Work}
\label{sec:related}

\textbf{Foundation Vision Encoders.}
Modern vision foundation models (VFMs) leverage diverse pretraining objectives to capture specific image properties. Self-supervised models such as MAE~\cite{mae}, DINOv1~\cite{caron2021emerging}, and DINOv2~\cite{oquab2024dinov2} provide exceptional structural and geometric descriptors. The recently introduced 7B-parameter DINOv3~\cite{simeoni2025dinov3} further utilizes Gram anchoring to preserve dense feature locality during large-scale training. In parallel, contrastive models like CLIP~\cite{radford2021clip} and SigLIP 2~\cite{tschannen2025siglip} align visual features with language, though often at the cost of spatial granularity. Other approaches, such as AIMv2~\cite{fini2025multimodal}, introduce multimodal autoregressive objectives to unify these capabilities, while SILC~\cite{naeem2024silc} combines contrastive learning with local self-distillation. The Segment Anything Model (SAM)~\cite{kirillov2023sam} on the other hand achieves unprecedented zero-shot segmentation through training on massive segmentation datasets. Recent breakthroughs, such as the Perception Encoder (PEcore)~\cite{bolya2025perception}, challenge the notion that these objectives are mutually exclusive by demonstrating that high-quality general features exist within the intermediate layers of a single, contrastively-trained network. Further, PElang~\cite{bolya2025perception} extends this by language-aligning these internal features for multimodal LLMs. However, these encoders are typically experts in limited task domains, and their out-of-domain performance is below expectations. Our work EUPE addresses this by distilling knowledge from multiple expert teachers into a single, universal student encoder.

\textbf{Knowledge Distillation for Vision Encoders.}
Knowledge distillation (KD), originally proposed by Hinton et al.~\cite{hinton2015distilling}, provides a general framework for training a compact student model to mimic a larger teacher. This foundational concept has been extended by numerous single-teacher distillation variants. Teacher Assistant Knowledge Distillation (TAKD)~\cite{mirzadeh2020improved} bridges a large capacity gap between teacher and student by introducing an intermediate-sized ``teacher assistant'' model. Other works focus on distilling specific capabilities from powerful foundation models: EfficientSAM~\cite{xiong2024efficientsam} leverages masked image pretraining to distill the segmentation capabilities of SAM into a much smaller encoder, and PEspatial~\cite{bolya2025perception} distills the strong spatial features found in the intermediate layers of the Perception Encoder. Techniques have also been developed to preserve specific feature properties during distillation, such as the Gram anchoring method in DINOv3~\cite{simeoni2025dinov3}, which maintains the quality of dense, local features throughout training. Our work builds upon the simple yet effective principles but extends it to a multi-teacher setting. We intentionally keep the per-teacher distillation flow as simple as possible to focus on the challenges of combining knowledge from multiple, diverse experts into small and efficient student.


\textbf{Agglomerative and Multi-Teacher Methods.}
To benefit from multiple strong encoders simultaneously, some work has explored the theoretical underpinnings of combining knowledge from multiple sources. Formont et al.~\cite{formont2025learning} proposed a task-agnostic, information-theoretic framework for multi-teacher distillation based on a majority-vote objective, while Ramtoula et al.~\cite{ramtoula2025fantastic} provides a systematic probing framework (``ComBo'') to identify and combine the most task-relevant features from disparate foundation models. Another direction is multi-teacher distillation. UNIC~\cite{sariyildiz2024unic} introduced a ``ladder of projectors'' and ``teacher dropping'' to prevent any single teacher from dominating the gradient, while its successor DUNE~\cite{sariyildiz2025dune} successfully merges 2D vision and 3D perception teachers through heterogeneous co-distillation. AM-RADIO~\cite{ranzinger2024radio} introduced an agglomerative framework for multi-teacher distillation that creates a unified student from CLIP, DINOv2, and SAM by progressively merging similar image tokens in the network's deeper layers; its successor RADIOv2.5~\cite{heinrich2025radiov2} further addressed resolution mode shifts and teacher imbalance. 
However, when it comes to an efficient computing scenario like on edge devices, these methods are not competitive to domain experts. Our work EUPE discovers the keep missing part is scaling-up to a proxy model before direct scaling down from multiple teachers.
\section{Efficient Universal Perception Encoder}
\label{sec:eupe}

\subsection{Pipeline Overview}
\begin{figure}[t]
    \centering
    \includegraphics[width=\linewidth]{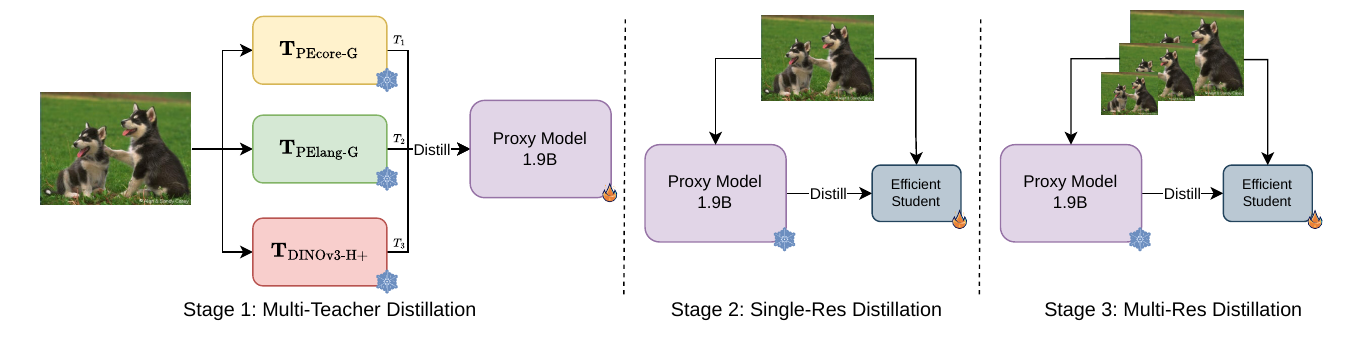}
    \caption{Multi-stage distillation pipeline (scaling up $\rightarrow$ scaling down). In Stage 1 we distill from multiple foundation models into a heavy proxy model. For Stage 2 the distillation happens from the proxy model into the target efficient encoder. In Stage 3 we finetune from Stage 2 models at multiple resolutions. The image pyramid indicates the multi-resolution inputs.}
    \label{fig:pipeline}
\end{figure}

We propose a multi-stage distillation pipeline with the principle: scaling up, then scaling, as shown in Fig. \ref{fig:pipeline}. We opt for simplicity in the design to demonstrate the importance of scaling up before scaling down. 

Our pipeline has three stages. The first stage is multi-teacher distillation into a large proxy model, e.g. 1.9B parameters. We select a heavy model as the proxy because it has enough capacity to learn universally good representations from diverse foundational encoders of different domains. The input consists of label-free images which are passed to all teachers in parallel at their native resolution. Each teacher outputs a class token and a set of patch tokens. The same image also passes through the proxy model and outputs a class token and patch tokens. They are compared with each teacher's tokens to compute the distillation loss. For teachers, we select representative foundation encoders from each task domain. PEcore \cite{bolya2025perception} is selected as the domain expert for zero-shot image classification and retrieval, and DINOv3 \cite{simeoni2025dinov3} is chosen as the domain expert for dense prediction. In addition, we find that PElang \cite{bolya2025perception} is crucial for vision-language modeling. 

The second stage is fixed-resolution distillation from the Stage 1 proxy model into the target efficient encoder. We hypothesize that it is much easier for efficient encoders to learn from a universal proxy teacher than directly learning from diverse domain experts, mainly because efficient encoders have lower capacity to effectively unify knowledge of multiple teachers into universal representations. In this stage, we keep the image resolution fixed at $256 \times 256$ so that the training step is computationally efficient and we can afford a longer learning schedule. 

The third stage is multi-resolution finetuning from the Stage 1 proxy model to the target efficient encoder. The student encoder is initialized from the Stage 2 distilled checkpoint. Instead of passing the same image to the teacher and the student, we resize the image several times into a pyramid and let the teacher and the student randomly select one scale from the pyramid independently. As a result, the student can learn from the teacher's representations at different granularity. 
This stage is designed to accommodate various resolutions of downstream tasks.

\begin{figure}[t]
    \centering
   \includegraphics[width=\linewidth]{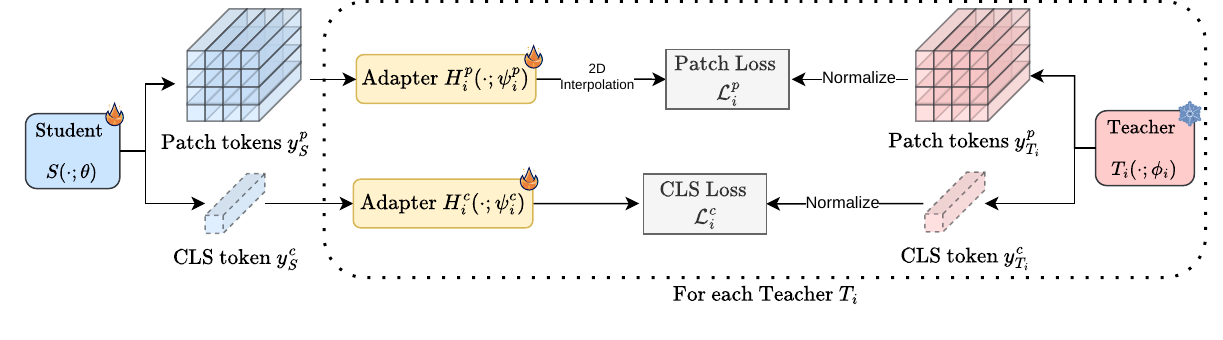}
    \caption{Per teacher distillation flow. Snowflake symbol indicates frozen parameters and flame symbol indicates trainable parameters. 2D interpolation is applied to the patch tokens in case the student's output and the teacher's output are of different spatial dimensions.}
    \label{fig:distill}
\end{figure}

In all stages, the distillation from the teacher to the student follows the same flow as in Fig. \ref{fig:distill}. Let $S(\cdot;\theta)$ be the student encoder parameterized by $\theta$, and $T_i(\cdot;\phi_i)$ be the $i^\text{th}$ teacher encoder parameterized by $\phi_i$ where $i$ can be greater than 1 in Stage 1. Given the student's input $x_S$ and the teacher's input $x_{T_i}$, they each output a class token $y_*^c$ and patch tokens $y_*^p$:
\begin{equation}
\bigl(y_S^c, y_S^p\bigr) = S(x_S;\theta),
\qquad y_S^c \in \mathbb{R}^{d_S},\; y_S^p \in \mathbb{R}^{N_S \times d_S}.
\end{equation}
\begin{equation}
\bigl(y_{T_i}^c, y_{T_i}^p\bigr) = T_i(x_{T_i};\phi_i),
\qquad y_{T_i}^c \in \mathbb{R}^{d_{T_i}},\; y_{T_i}^p \in \mathbb{R}^{N_{T_i} \times d_{T_i}}.
\end{equation}
where $d_S, N_S, d_{T_i}, N_{T_i}$ are the feature dimension and number of patch tokens for the student and the teacher, respectively. To connect the outputs of the student and the $i^\text{th}$ teacher, we append adapter head modules to the student outputs to match the feature dimensions. Specifically, let $H_i^c(\cdot;\psi_i^c)$ and $H_i^p(\cdot;\psi_i^p)$ be the adapter heads for the class token and patch tokens for the $i$th teacher parameterized by $\psi_i^c$ and $\psi_i^p$, respectively. Then the adapted class token and patch tokens for the $i^\text{th}$ teacher are:
\begin{equation}
\begin{aligned}
    z_{T_i}^c &= H_i^c(y_S^c;\psi_i^c), \qquad z_{T_i}^c \in \mathbb{R}^{d_{T_i}} \\
    z_{T_i}^p &= H_i^p(y_S^p;\psi_i^p), \qquad z_{T_i}^p \in \mathbb{R}^{N_S \times d_{T_i}}
\end{aligned}
\end{equation}
To match the spatial resolution between $z_{T_i}^p$ and $y_{T_i}^p$, we 2D-interpolate the smaller one into the larger size so they will have the same shape $\max(N_S, N_{T_i}) \times d_{T_i}$. Finally, the distillation loss $L_i$ between the student and the $i^\text{th}$ teacher is calculated from the student's adapted tokens $z_{T_i}^c, z_{T_i}^p$ and the teacher's normalized tokens $\bar{y}_{T_i}^c, \bar{y}_{T_i}^p$:
\begin{equation}
    L_i = L_i^c(z_{T_i}^c, \bar{y}_{T_i}^c) + L_i^p(z_{T_i}^p, \bar{y}_{T_i}^p)
\end{equation}
where $L_i^c, L_i^p$ are the class token loss and patch token loss, respectively, which we introduce below. During training, $\theta, \psi_i^c, \psi_i^p$ are learnable parameters and $\phi_i$ is frozen.

\subsection{Loss}
For simplicity, we use the same loss formulation for all stages. Following AM-RADIO \cite{ranzinger2024radio}, the class token loss is the cosine similarity loss and the patch token loss is a combination of the cosine similarity loss and the smooth L1 loss:
\begin{equation}
\begin{aligned}
    L_i^c(z_{T_i}^c, \bar{y}_{T_i}^c) &= L_{cos}(z_{T_i}^c, \bar{y}_{T_i}^c) \\
    L_i^p(z_{T_i}^p, \bar{y}_{T_i}^p) &= \alpha L_{cos}(z_{T_i}^p, \bar{y}_{T_i}^p) + \beta L_{smooth-L1}(z_{T_i}^p, \bar{y}_{T_i}^p)
\end{aligned}
\end{equation}
where $\alpha=0.9, \beta=0.1$ are the loss weights. The total distillation loss $L$ is the summation of the loss for all teachers:
\begin{equation}
    L = \sum_{i} L_i^c(z_{T_i}^c, \bar{y}_{T_i}^c) + L_i^p(z_{T_i}^p, \bar{y}_{T_i}^p)
\end{equation}
For Stage 1, $i$ ranges over the indices of all teachers. For Stage 2\&3, there is only one teacher (proxy model).

\subsection{Feature Normalization}
We normalize the teacher output during distillation, i.e., $y_{T_i}^c \rightarrow \bar{y}_{T_i}^c, y_{T_i}^p \rightarrow \bar{y}_{T_i}^p$. This helps stabilize the feature distillation \cite{heo2019comprehensive}, especially in Stage 1. As pointed out by UNIC \cite{sariyildiz2024unic}, the class token and patch tokens of a teacher's outputs can have very different feature mean norm and standard deviation. And these statistics across teachers are also very diverse. Distillation without feature normalization will cause the domination of one type of token (the class token in most cases) from a single teacher. Unlike the complex PHI-S normalization used in RADIOv2.5 \cite{heinrich2025radiov2}, we opt for simplicity by simply subtracting the mean and dividing by the standard deviation (std), which proves effective. We compute the normalization statistics by running each teacher through a tiny batch of the training data and then fix the mean and std for the rest of the training. This is also different from UNIC \cite{sariyildiz2024unic} which computes statistics on-the-fly during distillation using an exponential moving average. On-the-fly computation requires gathering the features across all GPUs every step, which consumes more memory and makes it hard to scale up the batch size on multiple nodes.

\subsection{Data}
For all stages, we train on the same DINOv3 dataset \cite{simeoni2025dinov3}, which consists of LVD-1689M with balanced coverage of all visual concepts appearing on
the Web and high quality public datasets such as ImageNet1k \cite{imagenet}. We also adopt the same data sampling strategy in DINOv3 to train with both homogeneous batches of data from ImageNet1k and heterogeneous batches from LVD-1689M. The probability of sampling from ImageNet1k is set to 10\%.

\section{Experiments}
\label{sec:exp}

In this section, we benchmark EUPE by comparing it to existing efficient vision encoders on a variety of computer vision tasks. To compare their generalization capability on multiple tasks, we keep \textit{all encoders frozen} and \textit{solely} use their representations without adapter heads. Our test bed consists of three mainstream vision task domains. One is image understanding to test the encoder's global representation, including zero-shot classification on ImageNet1k (IN1k-ZS) and KNN classification on ImageNet1k (IN1k-KNN). Another is dense prediction to measure their spatial understanding ability, including semantic segmentation (AKE20k \cite{ade}), monocular depth estimation (NYUv2 \cite{nyu}), and semantic keypoint correspondence estimation (SPair \cite{spair}). Finally we also test on the vision-language modeling tasks. We train a Llava \cite{llava} model with the encoder plugged in. We follow the definition proposed by Cambrian-1 \cite{tong2024cambrian} of four types of VLM benchmarks. We choose one or two representative benchmarks from each type, namely OCR (TextVQA \cite{textvqa}), knowledge (SQA \cite{sqa}), vision-centric (Realworld \cite{realworld} and POPE \cite{pope}), and general (GQA \cite{gqa} and MME \cite{mme}). We share more details on the setup in the supplementary material.

\subsection{Implementation Details}
In Stage 1, we choose the foundation teacher encoders to be PEcore-G (1.9B), PElang-G (1.7B), and DINOv3-H+ (840M). We follow the recipe described in the AM-Radio paper series. We run all teachers at their native resolutions (448 for PEcore/lang and 256 for DINOv3-H+) during training. We train a 1.9B parameter proxy model with 4 register tokens. We perform a crude centering of the teacher outputs by measuring their per-coordinate mean and variance during 500 iterations before training. We use the standard ImageNet constants for the mean-std normalization of inputs.

In Stage 2, we train with a $256\times256$ fixed resolution, a batch size of 8192, a cosine learning rate schedule, a base learning rate of $2\mathrm{e}{-5}$, and weight decay set to $1\mathrm{e}{-4}$ for 390k iterations. We augment the input images with random resized cropping, random horizontal flipping, color jittering, Gaussian blur, and random solarization. For efficient student encoders, we opt for backbones with less than 100M parameters. The ViT family includes ViT-B (86M), ViT-S (21M), and ViT-T (6M). The CNN family includes ConvNext-Base (89M), ConvNext-Small (50M), and ConvNext-Tiny (29M). 

In Stage 3, we build the image pyramid with three scales, i.e. 256, 384, and 512. All other data augmentation steps are the same as in Stage 2. The student and the teacher randomly select one scale from the pyramid independently for each iteration. We opt for a shorter learning schedule in finetuning with batch size of 4096, base learning rate of $1\mathrm{e}{-5}$ for 100k iterations.

For all adapter heads, we adopt a simple 2-layer MLP design which starts with a linear projection without bias, followed by LayerNorm and GELU, and ends with another linear projection without bias. The hidden dimension is 1536 in Stage 1 and 3072 in Stage 2\&3. Wherever spatial alignment is needed, we use PyTorch's builtin interpolation with bicubic mode to resize the patch tokens.

\subsection{Comparison with SOTA}

\begin{table*}[t]
\centering
\caption{Comparison with representative domain experts and agglomerative encoders across image understanding, VLM, and dense prediction benchmarks. Best results are indicated in \textbf{bold}. Numbers in brackets indicate the gap with the best domain expert. ``no txt'' means no text encoder. ``no cls'' means no class token output. $^*$The discrepancy with results from \cite{sariyildiz2025dune} is due to benchmarking only the encoder part without adapter head.}
\resizebox{\textwidth}{!}{%
\small
\setlength{\tabcolsep}{4pt}
\renewcommand{\arraystretch}{1.15}
\begin{tabular}{l cc cccccc ccc}
\toprule
& \multicolumn{2}{c}{\textbf{Image under.}}
& \textbf{VLM OCR} & \textbf{VLM know.} & \multicolumn{2}{c}{\textbf{VLM vision}} & \multicolumn{2}{c}{\textbf{VLM general}}
& \multicolumn{3}{c}{\textbf{Dense prediction}} \\

\cmidrule(lr){2-3}
\cmidrule(lr){4-9}
\cmidrule(lr){10-12}

\textbf{Model} 
& \rotatebox{90}{IN1k-ZS}
& \rotatebox{90}{IN1k-KNN}

& \rotatebox{90}{TextVQA}
& \rotatebox{90}{SQA}
& \rotatebox{90}{Realworld}
& \rotatebox{90}{POPE}
& \rotatebox{90}{GQA}
& \rotatebox{90}{MMEp}

& \rotatebox{90}{SPair}
& \rotatebox{90}{NYUv2$\downarrow$}
& \rotatebox{90}{ADE20k} \\

\midrule

\multicolumn{12}{l}{\textit{Domain Experts}} \\

PEcore-B \cite{bolya2025perception}
& 78.4 & 79.7
& 50.8 & \textbf{70.0} & 52.9 & 85.8 & 65.6 & 1375.5
& 25.9 & 0.641 & 37.4 \\

PEspatial-B \cite{bolya2025perception}
& no cls & no cls
& 42.8 & 69.0 & 51.6 & 84.4 & 63.3 & 1279.6
& 42.2 & 0.389 & 45.5 \\

SigLIP2-B \cite{tschannen2025siglip}
& 78.2 & 83.2
& \textbf{51.6} & 69.8 & 52.5 & 85.0 & 65.2 & \textbf{1389.5}
& 32.8 & 0.512 & 41.6 \\

DINOv3-ViT-B \cite{simeoni2025dinov3}
& no txt & 83.0
& 42.7 & 69.3 & 52.6 & 85.7 & 65.9 & 1368.0
& \textbf{51.3} & \textbf{0.373} & 51.8 \\

\midrule

\multicolumn{12}{l}{\textit{Agglomerative Methods}} \\

RADIOv2.5-B \cite{heinrich2025radiov2}
& 74.6 & 81.9
& 47.0 & 69.3 & 54.3 & 84.7 & 65.8 & 1349.8
& 48.7 & 0.435 & 49.0 \\

DUNE-B \cite{sariyildiz2025dune}
& no txt & 42.5
& 41.7 & 69.2 & 52.0 & 84.5 & 64.1 & 1294.8
& 39.4 & 0.375 & 40.9$^*$ \\


\rowcolor{gray!10}
\textbf{EUPE-ViT-B (Ours)}
& \textbf{79.7} & \textbf{84.1}
& 50.4 (1.2) & 69.7 (0.3) & \textbf{55.5} & \textbf{85.9} & \textbf{67.3} & 1374.5 (15.0)
& \textbf{51.3} & 0.391 (0.018) & \textbf{52.4} \\

\bottomrule
\end{tabular}
}
\label{tab:sota}
\end{table*}

We compare our model with both SOTA domain experts and previous agglomerative encoders on our test bed. We focus on efficient architectures and identify that the most common efficient backbone for all methods is ViT-B. We benchmark our EUPE-ViT-B and others, and report the performance in Table \ref{tab:sota} and Fig. \ref{fig:teaser}.

Overall, our EUPE-ViT-B is the most universally transferable encoder with on-par or even better performance on each benchmark across image understanding, dense prediction, and vision-language modeling when compared to the strongest model for that benchmark. 
Compared to agglomerative methods, it outperforms RADIOv2.5-B and DUNE-B on all VLM tasks and most dense prediction tasks by significant margins with only a small gap with DUNE-B on NYUv2. 
Compared to domain experts, it excels at image understanding on ImageNet1k, outperforms the dense prediction expert (DINOv3-ViT-B) on ADE20k, and outperforms the VLM experts (SigLIP2-B and PEcore-B) on Realworld, POPE, and GQA. On other benchmarks such as NYUv2, SQA, TextVQA, and MMEp, its gap with the corresponding domain expert is marginal. 

\subsection{Ablation Studies}
We detail our key ablation studies below. Further experiments regarding the data-mix, loss weights, and proxy model size are provided in the supplementary material. Unless otherwise specified, all ablations use the ViT-B architecture.

\begin{table*}[t]
\centering
\caption{Ablation on the necessity of the three-stage pipeline. ``Stage 2 only'' means direct distillation from multiple teachers into the target efficient encoder. Best results are indicated in \textbf{bold}.}
\small
\setlength{\tabcolsep}{10pt} 
\renewcommand{\arraystretch}{1.2}
\begin{tabular}{ll cccc}
\toprule
\textbf{Task domain} & \textbf{Benchmark} & \textbf{Stage 2 only} & \textbf{Stage 1\&2} & \textbf{Stage 1\&3} & \textbf{Stage 1\&2\&3} \\
\midrule
\multirow{2}{*}{Image} & IN1k-ZS & 79.6 & 79.5 & \textbf{80.0} & 79.7 \\
 & IN1k-KNN & 84.0 & 84.0 & \textbf{84.3} & 84.1 \\
\midrule
\multirow{6}{*}{VLM} & TextVQA & 46.8 & 48.3 & 49.5 & \textbf{50.4} \\
 & SQA & 69.6 & 69.3 & 69.2 & \textbf{69.7} \\
 & Realworld & 53.5 & 55.1 & 55.1 & \textbf{55.5} \\
 & POPE & 85.3 & 85.3 & 84.6 & \textbf{85.9} \\
 & GQA & 66.6 & 66.4 & \textbf{67.3} & \textbf{67.3} \\
 & MMEp & 1337.9 & 1345.6 & \textbf{1399.7} & 1374.5 \\
\midrule
\multirow{3}{*}{Dense} & SPair & 35.1 & 41.0 & \textbf{53.3} & 51.3 \\
 & NYUv2$\downarrow$ & 0.616 & 0.557 & \textbf{0.388} & 0.391 \\
 & ADE20k & 41.9 & 43.3 & 52.0 & \textbf{52.4} \\
\bottomrule
\end{tabular}%
\label{tab:stages}
\end{table*}

\textbf{Necessity of stages.} Table~\ref{tab:stages} validates that all three stages contribute complementary gains.
Using only Stage~2 (direct multi-teacher distillation into efficient student) yields weaker VLM performance, especially on OCR, and also poor dense prediction performance.
Adding Stage~1 significantly improves vision-language modeling tasks such as TextVQA and Realworld, but this setup still lags behind the full pipeline on dense tasks.
The Stage 1+3 variant performs multi-resolution distillation after Stage 1. In this case, we adopt the same learning schedule as in Stage 2. This setting gives the strongest performance on dense prediction tasks, e.g. SPair (53.3) and NYUv2 (0.388), but the gaps behind the domain experts for VLM are significant. Also, training with multi-resolution is computationally costly, and we cannot afford a long schedule. The time to run one iteration in Stage 3 is twice as long as in Stage 2. Therefore, we opt for a long fixed-resolution training in Stage 2 followed by a short multi-resolution training in Stage 3. This setting improves the VLM metrics in general without sacrificing image and dense performance too much, resulting in the best overall balance.


\begin{table*}[t]
\centering
\caption{Ablation on the choice of teacher foundation encoders in Stage 1. SOTA is the best per-benchmark performance among all existing vision encoders as a reference. PEc = PEcore-G. PEl = PElang-G. S2 = SigLIP2-G. Dv3 = DINOv3-H+. Best results are in \textbf{bold}.}
\small
\setlength{\tabcolsep}{12pt} 
\renewcommand{\arraystretch}{1.2}
\begin{tabular}{ll cccc}
\toprule
\textbf{Task domain} & \textbf{Benchmark} & \textbf{SOTA} & \textbf{PEc\&Dv3} & \textbf{PEc\&Dv3\&S2} & \textbf{PEc\&Dv3\&PEl} \\
\midrule
\multirow{2}{*}{Image} & IN1k-ZS & 78.4 & \textbf{79.9} & 78.8 & 79.7 \\
 & IN1k-KNN & 83.2 & \textbf{84.3} & 84.2 & 84.1 \\
\midrule
\multirow{6}{*}{VLM} & TextVQA & \textbf{51.6} & 48.6 & 44.8 & 50.4 \\
 & SQA & 70.0 & 69.7 & \textbf{70.2} & 69.7 \\
 & Realworld & 54.3 & 55.1 & 52.9 & \textbf{55.5} \\
 & POPE & 85.8 & 85.8 & 84.7 & \textbf{85.9} \\
 & GQA & 65.9 & 66.7 & 66.4 & \textbf{67.3} \\
 & MMEp & \textbf{1389.5} & 1375.2 & 1271.6 & 1374.5 \\
\midrule
\multirow{3}{*}{Dense} & SPair-71k & 51.3 & 51.5 & \textbf{52.1} & 51.3 \\
 & NYUv2$\downarrow$ & \textbf{0.373} & 0.384 & 0.401 & 0.391 \\
 & ADE20k & 51.8 & \textbf{52.5} & \textbf{52.5} & 52.4 \\
\bottomrule
\end{tabular}%
\label{tab:teachers}
\end{table*}

\begin{table*}[t]
\centering
\caption{Proxy model performance with different teachers sets used to train the Stage-1 proxy. Also include teachers (PEcore-G, DINOv3-H+) performance as a reference. PEc = PEcore-G. PEl = PElang-G. S2 = SigLIP2-G. Dv3 = DINOv3-H+. Best results are in \textbf{bold}.}
\small
\setlength{\tabcolsep}{10pt} 
\renewcommand{\arraystretch}{1.2}
\begin{tabular}{ll ccccc}
\toprule
\textbf{Task domain} & \textbf{Benchmark} & \textbf{Dv3} & \textbf{PEc} & \textbf{PEc\&Dv3} & \textbf{PEc\&Dv3\&S2} & \textbf{PEc\&Dv3\&PEl}\\
\midrule
\multirow{2}{*}{Image} & IN1k-ZS & no text & \textbf{85.4}& 85.0 & 85.3 & 84.8 \\
 & IN1k-KNN & 85.4 & \textbf{87.2} & 87.0 & \textbf{87.2} & 87.0 \\
\midrule
\multirow{6}{*}{VLM} & TextVQA & 49.8 & 54.7 & 56.2 & 53.2 & \textbf{58.6} \\
 & SQA & 69.0 & \textbf{72.2} & 70.9 & 70.2 & 70.6 \\
 & Realworld & 53.9 & 56.9 & 59.24 & 54.2 & \textbf{60.4} \\
 & POPE & 87.2 & 85.9 & \textbf{87.3} & 87.0 & \textbf{87.3} \\
 & GQA & 67.9 & 67.0 & 68.6 & 68.4 & \textbf{69.2} \\
 & MMEp & 1385.0 & \textbf{1456.6} & 1455.4 & 1366.2 & 1450.2 \\
\midrule
\multirow{3}{*}{Dense} & SPair-71k & 49.7 & 20.3 & 52.9 & \textbf{54.4} & 53.8 \\
 & NYUv2$\downarrow$ & 0.352 & 0.590 & 0.332& \textbf{0.321} & 0.390  \\
 & ADE20k & 54.8 & 38.7 & \textbf{56.0} & 55.4 & 55.9 \\
\bottomrule
\end{tabular}%
\label{tab:proxy_perf}
\end{table*}

\textbf{The choice of teacher foundation models.} Table~\ref{tab:teachers} shows that selecting the right combination of teacher models in Stage 1 matters. The teacher set affects which capabilities are emphasized. We start with combining PEcore-G and DINOv3-H+, which shows promising signals on image understanding and dense prediction tasks. However, the gap with SOTA performance on the VLM OCR benchmark is huge. 
Then we explore adding another strong expert on VLM OCR tasks. SigLIP2-G itself achieves superior performance on all VLM benchmarks, but it substantially degrades the OCR metric when combined with PEcore-G and DINOv3-H+. This indicates that SigLIP2 features may not be compatible with the other teachers. Our hypothesis is that it is not helpful to have two CLIP-style models like PEcore-G and SigLIP2-G in the combination at the same time. PElang-G is a language-focused model derived from PEcore-G through alignment with language models. It turns out to be a good complement to the combinations. 
Adding PElang-G provides the strongest OCR and general VLM performance among the compared sets, without sacrificing the image and dense performance too much.
We therefore use PEcore-G, PElang-G, and DINOv3-H+ as the default set to maximize multi-task robustness.


\textbf{Performance of proxy models.} We also report the performance of different proxy models as a reference. Table \ref{tab:proxy_perf} shows that PEcore-G and DINOv3-H+ are experts in VLM and dense prediction, respectively. Combining them together provides a good foundation for all three task domains. The PElang-G is crucial for VLM tasks especially OCR-related. SigLIP2-G, on the other hand, does not work well with PEcore-G and DINOv3-H+, causing major degradations in VLM performance. These observations align with the final results of targeted efficient encoder in Table \ref{tab:teachers}, indicating that the student learns well from the teacher. 

\subsection{Feature Visualization}

\begin{figure}[t!]
    \centering
    \includegraphics[width=0.8\textwidth]{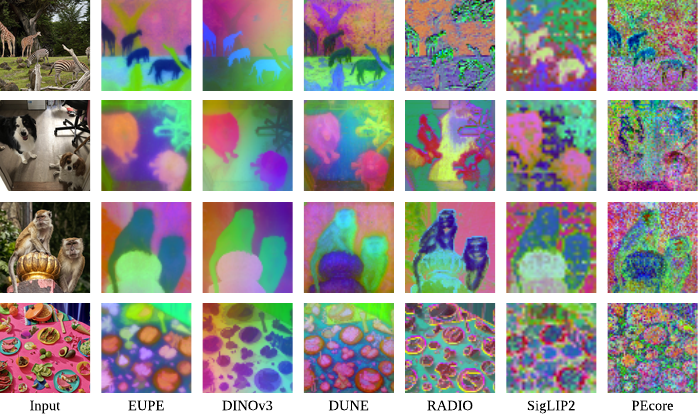}
    \caption{Comparison of dense features by projecting the patch tokens using PCA into RGB space. From left to right: EUPE-ViT-B16 (Ours), DINOv3-ViT-B16, DUNE-B14, RADIOv2.5-B16, SigLIP2-B16, PEcore-B16. Best viewed in color.}
    \label{fig:vis}
\end{figure}

To qualitatively compare the model's feature representations, we project dense patch tokens into a three-dimensional space using Principal Component Analysis (PCA) and map these dimensions to RGB. We apply this visualization technique to both domain expert and agglomerative ViT-Bs, as illustrated in Fig. \ref{fig:vis}.

For models trained with image-text pairs like PEcore and SigLIP2, their patch tokens contain semantic information but are not spatially consistent, leading to noisy representations. DINOv3, on the other hand, has highly sharp features with semantic coherence, but lacks discrimination ability for fine-grained details (e.g. food and plates having similar representations) as shown in the last row. For the agglomerative DUNE model, the features are similar to DINOv3 due to distilling from multiple dense prediction experts. Our EUPE model can combine the best of both worlds, i.e. achieving both \textit{semantically sharp features} and \textit{sensitivity to fine-grained details}. For the other agglomerative model, RADIO, its features are overly sensitive, which breaks the semantic coherence (e.g., in row 2, the black fur of the dogs merges with the background).

\begin{figure}[t]
    \centering
    \includegraphics[width=0.95\textwidth]{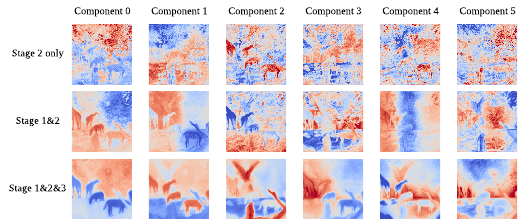}
    \caption{Comparison of dense features PCA components for encoders trained with stage variants. Input is the same as the image in Fig. \ref{fig:vis} row 1. ``Stage 2 only'' means direct distillation from multiple teachers into the target efficient encoder. Best viewed in color.}
    \label{fig:pca}
\end{figure}

We further analyze the features of encoders trained with different stage settings. Fig. \ref{fig:pca} helps us to understand their difference through visualization of the first few principal components. The encoder trained with Stage 2 only shows noisy feature maps and it is hard to identify semantic coherence. This can be effectively addressed by scaling up in Stage 1 and then scaling down as shown in the second row, indicating that learning from a single universal large teacher is an easier path compared to directly learning from multiple domain experts for efficient encoders. However, without Stage 3 multi-resolution training, the semantic coherence can be broken by spatial awareness. As shown in components 1\&4 of row 2, the visualization is divided into local regions due to resolution mismatch during training and inference. Stage 3 training can address this issue and makes the feature representations even sharper.

\subsection{Full Family of EUPE}

The full family includes variants based on the Vision Transformer (ViT) and ConvNeXt architectures. These models cover a wide range of parameter sizes and inference costs to accommodate diverse on-device use cases under different computation budgets. In addition to the results in Table \ref{tab:sota}, Table \ref{tab:vits} and \ref{tab:cnns} reports the comparison of the remaining EUPE family versus other model collections of the corresponding size.

\begin{table*}[t]
\centering
\caption{ViT family evaluation across image understanding, VLM, and dense prediction tasks. ``no cls'' means no class token output. ``no txt'' means no text encoder.}
\resizebox{\textwidth}{!}{%
\small
\setlength{\tabcolsep}{3pt}
\renewcommand{\arraystretch}{1.12}
\begin{tabular}{l cc cccccc ccc}
\toprule
\multirow{2}{*}{\textbf{Model}} & \multicolumn{2}{c}{\textbf{Image Under.}} & \multicolumn{1}{c}{\textbf{VLM OCR}} & \multicolumn{1}{c}{\textbf{VLM Know.}} & \multicolumn{2}{c}{\textbf{VLM Vision}} & \multicolumn{2}{c}{\textbf{VLM General}} & \multicolumn{3}{c}{\textbf{Dense Prediction}} \\
\cmidrule(lr){2-3} \cmidrule(lr){4-4} \cmidrule(lr){5-5} \cmidrule(lr){6-7} \cmidrule(lr){8-9} \cmidrule(lr){10-12}
& IN1k-ZS & IN1k-KNN & TextVQA & SQA & Realworld & POPE & GQA & MMEp & SPair & NYUv2$\downarrow$ & ADE20k \\
\midrule
PEcore-T       & 52.9 & 61.7 & 44.3 & 68.6 & 47.4 & 82.2 & 60.9 & 1221.1 & 20.2 & 0.756 & 22.0 \\
PEspatial-T    & no cls & no cls & 40.7 & 68.0 & 48.7 & 80.7 & 59.0 & 1128.0 & 18.4 & 0.589 & 28.5 \\
\rowcolor{gray!10}
\textbf{EUPE-ViT-T}     & 50.5 & 66.3 & 42.0 & 69.5 & 50.0 & 82.4 & 61.4 & 1258.0 & 37.2 & 0.571 & 36.7 \\
\midrule
PEcore-S       & 62.8 & 71.9 & 47.0 & 69.4 & 48.4 & 83.2 & 63.5 & 1339.2 & 28.6 & 0.599 & 33.7 \\
PEspatial-S    & no cls & no cls & 42.5 & 67.6 & 49.0 & 82.8 & 62.9 & 1245.4 & 30.8 & 0.464 & 38.6 \\
DINOv3-ViT-S   & no txt & 78.6 & 41.7 & 68.5 & 50.3 & 84.5 & 64.2 & 1263.1 & 47.9 & 0.404 & 46.9 \\
\rowcolor{gray!10}
\textbf{EUPE-ViT-S}     & 69.8 & 78.2 & 44.1 & 69.3 & 51.7 & 84.5 & 65.0 & 1304.9 & 46.5 & 0.455 & 46.6 \\
\bottomrule
\end{tabular}%
}
\label{tab:vits}
\end{table*}

\begin{table*}[t]
\centering
\caption{ConvNext family evaluation across VLM and dense prediction tasks. We omit results on image understanding tasks as the models do not have class token output.}
\resizebox{\textwidth}{!}{%
\small
\setlength{\tabcolsep}{4pt}
\renewcommand{\arraystretch}{1.12}
\begin{tabular}{l cccccc ccc}
\toprule
\multirow{2}{*}{\textbf{Model}} & \multicolumn{1}{c}{\textbf{VLM OCR}} & \multicolumn{1}{c}{\textbf{VLM Know.}} & \multicolumn{2}{c}{\textbf{VLM Vision}} & \multicolumn{2}{c}{\textbf{VLM General}} & \multicolumn{3}{c}{\textbf{Dense Prediction}} \\
\cmidrule(lr){2-2} \cmidrule(lr){3-3} \cmidrule(lr){4-5} \cmidrule(lr){6-7} \cmidrule(lr){8-10}
 & TextVQA & SQA & Realworld & POPE & GQA & MMEp & SPair & NYUv2$\downarrow$ & ADE20k \\
\midrule
DINOv3-ConvNext-T & 41.6 & 69.8 & 44.7 & 81.9 & 60.3 & 1226.0 & 35.7 & 0.448 & 42.7 \\
\rowcolor{gray!10}
\textbf{EUPE-ConvNext-T} & 43.7 & 68.8 & 47.9 & 83.4 & 63.0 & 1278.1 & 41.3 & 0.430 & 43.5 \\
\midrule
DINOv3-ConvNext-S & 42.6 & 68.8 & 49.3 & 83.7 & 63.1 & 1321.5 & 34.7 & 0.432 & 44.8 \\
\rowcolor{gray!10}
\textbf{EUPE-ConvNext-S} & 45.0 & 68.9 & 50.5 & 84.0 & 64.7 & 1284.2 & 40.1 & 0.388 & 46.8 \\
\midrule
DINOv3-ConvNext-B & 42.7 & 69.1 & 46.6 & 84.4 & 63.7 & 1278.8 & 35.0 & 0.420 & 46.3 \\
\rowcolor{gray!10}
\textbf{EUPE-ConvNext-B} & 46.4 & 70.1 & 53.3 & 84.7 & 65.8 & 1348.9 & 37.7 & 0.365 & 48.9 \\
\bottomrule
\end{tabular}%
}
\label{tab:cnns}
\end{table*}

For the ViT-S/T family in Table \ref{tab:vits}, our EUPE models offer balanced performance across three task domains. At Tiny scale, EUPE achieves large margins on dense prediction tasks. At Small scale, EUPE approaches DINOv3-level performance on SPair and ADE20k, while preserving or even improving vision-language modeling capability over PEcore.

For the ConvNext family in Table \ref{tab:cnns}, our EUPE family consistently performs better compared to the domain expert DINOv3 family across Tiny, Small and Base sizes on dense prediction tasks. Additionally, EUPE unlocks better vision-language modeling capability, especially for the OCR and vision-centric cases.


\section{Conclusion}
\label{sec:conclusion}

We introduced EUPE, a simple yet effective recipe to obtain \emph{efficient universal perception encoders} by first \emph{scaling up} knowledge aggregation and then \emph{scaling down} to compact student models.
Across diverse benchmarks spanning image understanding, vision-language modeling, and dense prediction, EUPE yields balanced zero-shot transfer and consistently strong performance with little to no task-specific finetuning.
We hope EUPE serves as a practical foundation for deploying versatile vision systems under tight computational budgets, and as a baseline for future work on scaling proxy teachers and improving universal representations for edge and multi-task settings.

\clearpage
\newpage

\section*{Acknowledgements}
We thank Bilge Soran for leadership support. Thank Daniel Bolya, Christoph Feichtenhofer, and the broader Perception Team for making the PE model available. Thank Hu Xu and Daniel Li for sharing the MetaCLIP data.

\beginappendix
\section{Additional Ablation Studies}
In this section, we provide further experiments regarding the proxy model size, datamix, and loss weights.

\begin{table*}[h]
\centering
\caption{Proxy model performance with different teachers sets used to train the Stage-1 proxy. PEc = PEcore-G. PEl = PElang-G. S2 = SigLIP2-G. Dv3 = DINOv3-H+. Dv3-7B = DINOv3-7B. The number in brackets is the parameter size of the proxy model. Best results are in \textbf{bold}.}
\small
\setlength{\tabcolsep}{6pt} 
\renewcommand{\arraystretch}{1.2}
\begin{tabular}{ll cccc}
\toprule
& & \textbf{PEc\&Dv3} & \textbf{PEc\&Dv3\&S2} & \textbf{PEc\&Dv3\&PEl} & \textbf{PEc\&Dv3-7B\&PEl}\\
\textbf{Task domain} & \textbf{Benchmark} & \textbf{(1.9B)} & \textbf{(1.9B)} & \textbf{(1.9B)} & \textbf{(7B)} \\
\midrule
\multirow{2}{*}{Image} & IN1k-ZS & 85.0 & \textbf{85.3} & 84.8  & 85.1 \\
 & IN1k-KNN & 87.0 & \textbf{87.2} & 87.0 & \textbf{87.2} \\
\midrule
\multirow{6}{*}{VLM} & TextVQA & 56.2 & 53.2 & 58.6 & \textbf{59.7} \\
 & SQA & 70.9 & 70.2 & 70.6 & \textbf{71.6} \\
 & Realworld & 59.24 & 54.2 & \textbf{60.4} & 60.2 \\
 & POPE & \textbf{87.3} & 87.0 & \textbf{87.3} & 86.6 \\
 & GQA & 68.6 & 68.4 & 69.2 & \textbf{69.4}\\
 & MMEp & 1455.4 & 1366.2 & 1450.2 & \textbf{1484.9} \\
\midrule
\multirow{3}{*}{Dense} & SPair & 52.9 & 54.4 & 53.8 &  \textbf{56.2} \\
 & NYUv2$\downarrow$ & 0.332& 0.321 & 0.390 & \textbf{0.305} \\
 & ADE20k & 56.0 & 55.4 & 55.9 & \textbf{56.9} \\
\bottomrule
\end{tabular}%
\label{tab:proxy_7B}
\end{table*}

\noindent \textbf{Impact of scaling up the teachers.} 
We explore whether further scaling up the teachers can keep increasing performance. In the main paper setting, we used DINOv3-H+ (840M) in Stage 1 and the proxy model is ViT-G (1.8B) in Stage 2\&3. Here we simultaneously scale up the DINOv3 teacher in Stage 1 and the proxy model in Stage 2\&3 to ViT-7B scale. 

\begin{table*}[t]
\centering
\small
\setlength{\tabcolsep}{6pt}
\caption{Impact of scaling up the DINOv3 teacher in Stage 1 and the proxy model in Stage 2\&3. Reported performance is from the final ViT-B student after Stage 3.}
\resizebox{\textwidth}{!}{%
\begin{tabular}{ll cc cccccc ccc}
\toprule
& & \multicolumn{2}{c}{\textbf{Image under.}} & \multicolumn{1}{c}{\textbf{VLM OCR}} & \multicolumn{1}{c}{\textbf{VLM know.}} & \multicolumn{2}{c}{\textbf{VLM vision}} & \multicolumn{2}{c}{\textbf{VLM general}} & \multicolumn{3}{c}{\textbf{Dense prediction}} \\
\cmidrule(lr){3-4} \cmidrule(lr){5-10} \cmidrule(lr){11-13}
 \textbf{DINOv3 size} & \textbf{Proxy size} 
 & \rotatebox{90}{IN1k-ZS}
 & \rotatebox{90}{IN1k-KNN}
 & \rotatebox{90}{TextVQA}
 & \rotatebox{90}{SQA}
 & \rotatebox{90}{Realworld}
 & \rotatebox{90}{POPE}
 & \rotatebox{90}{GQA}
 & \rotatebox{90}{MMEp}
 & \rotatebox{90}{SPair}
 & \rotatebox{90}{NYUv2$\downarrow$}
 & \rotatebox{90}{ADE20k} \\
\midrule
ViT-H+ & ViT-G & 79.7 & 84.1 & 50.4 & 69.7 & 55.5 & 85.9 & 67.3 & 1374.5 & 51.3 & 0.391 & 52.4 \\
ViT-7B & ViT-7B & 80.2 & 83.9 & 48.5 & 69.8 & 53.9 & 85.3 & 66.6 & 1345.2 & 52.0 & 0.390 & 52.5 \\
\bottomrule
\end{tabular}%
}
\label{tab:proxy_scaling}
\end{table*}

Table \ref{tab:proxy_7B} verifies that scaling both the DINOv3 teacher and the proxy model to 7B can set new records on most benchmarks compared to the existing 1.9B proxy models. And on IN1k-ZS, Realworld, and POPE, the performance gap to the best proxy model is marginal. This is promising, but when distilling this 7B proxy model into the ViT-B student, we observe mixed signals as shown in Table \ref{tab:proxy_scaling}. Although image understanding and dense prediction have been slightly improved, the VLM quality is generally worse than before. Almost all benchmarks in VLM are dropped and major degradations are observed on TextVQA, Realworld, and MMEp. This indicates that the proxy's knowledge is not fully distilled to the student. The main reason could be the huge size difference between the 7B proxy and the 86M ViT-B. A possible solution may be progressive distillation through the Teaching Assistant proposed in \cite{mirzadeh2020improved}, which we leave as future work.

\noindent \textbf{Impact of datamix.}
We also compare the effect of training with LVD-1689M \cite{simeoni2025dinov3} and MetaCLIP \cite{metaclip}. We keep the probability of sampling from ImageNet1k the same as 10\% and vary the heterogeneous batches between LVD and MetaCLIP. The teachers in Stage 1 are PEcore-G and DINOv3-H+. The proxy model in Stage 2\&3 is 1.9B. Table \ref{tab:datamix} shows that despite the fact that MetaCLIP has 2.5B images, about 0.8B more than LVD, training on LVD yields better performance on almost all benchmarks, indicating the higher quality of LVD. 

\begin{table*}[t]
\centering
\small
\setlength{\tabcolsep}{6pt}
\caption{Impact of training on LVD-1689M with 1689M images versus MetaCLIP with 2.5B images}
\resizebox{\textwidth}{!}{%
\begin{tabular}{l cc cccccc ccc}
\toprule
 & \multicolumn{2}{c}{\textbf{Image under.}} & \multicolumn{1}{c}{\textbf{VLM OCR}} & \multicolumn{1}{c}{\textbf{VLM know.}} & \multicolumn{2}{c}{\textbf{VLM vision}} & \multicolumn{2}{c}{\textbf{VLM general}} & \multicolumn{3}{c}{\textbf{Dense prediction}} \\
\cmidrule(lr){2-3} \cmidrule(lr){4-9} \cmidrule(lr){10-12}
 \textbf{Training data} & \rotatebox{90}{IN1k-ZS}
 & \rotatebox{90}{IN1k-KNN}
 & \rotatebox{90}{TextVQA}
 & \rotatebox{90}{SQA}
 & \rotatebox{90}{Realworld}
 & \rotatebox{90}{POPE}
 & \rotatebox{90}{GQA}
 & \rotatebox{90}{MMEp}
 & \rotatebox{90}{SPair}
 & \rotatebox{90}{NYUv2$\downarrow$}
 & \rotatebox{90}{ADE20k} \\
\midrule
90\% MetaCLIP + 10\% IN1k & 79.3 & 83.7 & 48.5 & 69.7 & 54.2 & 83.9 & 66.8 & 1327.8 & 49.0 & 0.393 & 52.6 \\
90\% LVD + 10\% IN1k & 79.9 & 84.3 & 48.6 & 69.7 & 55.1 & 85.8 & 66.7 & 1375.2 & 51.5 & 0.384 & 52.5 \\
\bottomrule
\end{tabular}%
}
\label{tab:datamix}
\end{table*}

\noindent \textbf{Impact of varying patch loss weight}
In the early exploration of this work, we observed that the patch loss of DINOv3 teacher behaves differently from other teachers during distillation, therefore ablating it with different weights. We introduce a hyperparameter $\gamma$ in the distillation loss of DINOv3 teacher $L_{Dv3}$:
\begin{equation}
    L_{Dv3} = L^c(z_{Dv3}^c, \bar{y}_{Dv3}^c) + \gamma L^p(z_{Dv3}^p, \bar{y}_{Dv3}^p)
\label{eq:gamma}
\end{equation}
And it contributes to the total loss in the same way as other teachers shown in Eq. (6) in the main paper. We adopt the ``Stage 2 only'' setting by directly distilling multiple teachers into a ViT-B student, with teachers including PEcore-G, PElang-G and DINOv3-H+. Table \ref{tab:patch_loss} reports the results. In general, a higher patch loss weight ($\gamma=2.0$) gives better image understanding and dense prediction results, but leads to worse vision-language modeling on TextVQA, SQA, and Realworld. On the other hand, ignoring DINOv3's patch tokens ($\gamma=0.0$) leads to poor dense prediction performance despite a superior result on TextVQA. This means that DINOv3's patch tokens play an important role in dense prediction tasks but can hurt several VLM benchmarks if putting too much weight on them. To achieve a balanced performance, we keep the weight the same as the other teachers and transfer this setting to the training of the large proxy model in Stage 1.

\begin{table*}[t]
\centering
\setlength{\tabcolsep}{6pt}
\caption{Impact of varying the patch loss weight for DINOv3 in Eq. \ref{eq:gamma}. SPair@224 means the benchmark is done at $224\times224$ resolution. }
\renewcommand{\arraystretch}{1.15}
\begin{tabular}{l c cccccc cc}
\toprule
 & \textbf{Image} & \multicolumn{1}{c}{\textbf{VLM OCR}} & \multicolumn{1}{c}{\textbf{VLM know.}} & \multicolumn{2}{c}{\textbf{VLM vision}} & \multicolumn{2}{c}{\textbf{VLM general}} & \multicolumn{2}{c}{\textbf{Dense}} \\
\cmidrule(lr){2-2} \cmidrule(lr){3-8} \cmidrule(lr){9-10}
 \textbf{$\gamma$}
 & \rotatebox{90}{IN1k-KNN}
 & \rotatebox{90}{TextVQA}
 & \rotatebox{90}{SQA}
 & \rotatebox{90}{Realworld}
 & \rotatebox{90}{POPE}
 & \rotatebox{90}{GQA}
 & \rotatebox{90}{MMEp}
 & \rotatebox{90}{SPair@224}
 & \rotatebox{90}{ADE20k} \\
\midrule
0.0 & 79.7 & 51.8 & 71.4 & 54.3 & 84.9 & 66.4 & 1362.8 & 23.4 & 28.7 \\
1.0 & 80.2 & 50.9 & 69.9 & 55.2 & 84.9 & 66.4 & 1375.4 & 29.0 & 31.9 \\
2.0 & 80.3 & 50.1 & 68.8 & 54.0 & 86.2 & 66.6 & 1380.9 & 31.3 & 32.9 \\
\bottomrule
\end{tabular}%
\label{tab:patch_loss}
\end{table*}

\section{Inference Cost Comparison}
To power AI use cases on real edge devices, model inference cost is an important factor to take into consideration when down-selecting the most suitable architecture for the best user experience. Therefore, we provide both inference FLOPs and on-device latency for all models in our EUPE family in Table \ref{tab:cost}. The inference latency measurement is done by exporting the encoders as ExecuTorch models and profiling the models on mobile devices. We also report the cost of larger architectures not included in our EUPE family as a reference to show their limitation to be deployed on edge devices. 

\begin{table}[t]
\centering
\caption{Model size and inference cost comparison. We present per model the number of parameters and the cost measured by FLOPs and latency on images of size $256\times256$ and $512\times512$. Latency is measured on iPhone 15 Pro CPU. $^*$models not included in our EUPE family but to show their incompatibility for running on edge devices.}
\setlength{\tabcolsep}{8pt}
\renewcommand{\arraystretch}{1.15}
\begin{tabular}{lcccccc}
\toprule
 &  & \multicolumn{2}{c}{\textbf{Inference GFLOPs}} & \multicolumn{2}{c}{\textbf{Inference latency (ms)}} \\
\cmidrule(lr){3-4} \cmidrule(lr){5-6}
\textbf{Model} & \textbf{\#Params} & Res. 256 & Res. 512 & Res. 256 & Res. 512  \\
\midrule
ConvNext-Tiny  & 29M  & 5   & 20 & 22.4 & 82.4 \\
ConvNext-Small & 50M  & 11  & 46  & 38.5 & 141.9 \\
ConvNext-Base  & 89M  & 20  & 81  & 59.3 & 222.7 \\
ConvNext-Large$^*$ & 198M & 38  & 152 & 112.5 & 447.2 \\
\midrule
ViT-T     & 6M & 4 & 17 & 6.8 & 38.3 \\
ViT-S     & 21M  & 12  & 63  & 17.1 & 97.9 \\
ViT-B     & 86M  & 47  & 216  & 55.2 & 305.2 \\
ViT-L$^*$     & 300M & 163 & 721 & 192.6 & 990.4 \\
\bottomrule
\end{tabular}
\label{tab:cost}
\end{table}

When the model size is less than 100M parameters, we observe acceptable inference latency even with a resolution as high as 512. It is recommended to select ConvNext architectures at higher resolutions and ViT architectures for the low resolution scenario. Also note that small FLOPs of ConvNext do not necessarily lead to lower latency compared to ViT. This is because convolutional operations are often less efficient on CPU architecture compared to the highly optimized Matrix Multiplication (GEMM) operations used in ViTs.

\section{Detailed Benchmark Settings}
In this section, we provide details about the settings across all benchmarks in this paper, including the datasets, the additional training recipe if any, and the evaluation protocols. 

\subsection{Image Understanding}
We evaluate the global quality of vision encoders through image classification on the ImageNet1k \cite{imagenet} validation set and report the top-1 accuracy. For each image, we input at $224\times224$ resolution and take the class token of the vision encoder as the feature representation of that image. The class token is then used to predict the category label of the image using two protocols: 1) KNN; 2) zero-shot. In the KNN protocol, we pre-generate the class tokens on the images from the training set. Given the class token of a test image, we select $k$ images from the training set with the closest L2 distances between their class tokens and the test class token. Then the majority of their categories is chosen as the predicted label. We set $k=10$ in the KNN protocol. In the zero-shot protocol, we use the text tower of the vision encoder to build the zero-shot classifier weights with the text input being the 1000 category names of ImageNet1k. Given the class token of a test image, we compute the dot product of it and the classifier weights followed by softmax to output a probability distribution. The class name with the highest probability is the final prediction. Note that for our EUPE ViT family, we use the teacher's text tower to build the classifier weights and project the class token into the teacher's space using the adapter head.

\subsection{Vision-Language Modeling}
In this task domain, we evaluate the quality of patch tokens from the vision encoders by connecting them to a language model with an MLP projector following the LlaVA-1.5 \cite{llava} paradigm. We keep everything in LlaVA unchanged except swapping its vision encoder with the ones to be tested. We first train only the projector on 558K image-text pairs for vision-language alignment. Then we finetune both the projector and the language model on 665K language-image instruction-following data. For both stages, we train with consine learning rate schedule, 0.03 learing rate warmup ratio, 0 weight decay, AdamW as the optimizer for 1 epoch. The learning rate for the first stage and the second stage is $1e-3$ and $2e-5$, respectively. We use input resolution $336\times336$ for vision encoders with patch size 14 and $384\times384$ for vision encoders with patch size 16 to keep the number of visual tokens fixed. After the 2-stage training, we evaluate the model on 6 benchmarks from 4 types of tasks defined by Cambrian-1 \cite{tong2024cambrian}, i.e. TextVQA \cite{textvqa} for OCR, SQA \cite{sqa} for knowledge, Realworld \cite{realworld} and POPE \cite{pope} for vision-centric, and GQA \cite{gqa} and MME \cite{mme} for general. For POPE we report the F1 score. For MME we report its perception score. For all others, we report the accuracy.

\subsection{Dense Prediction: Semantic Segmentation}
We evaluate the performance of vision encoders in semantic segmentation using linear probing on the ADE20k dataset \cite{ade}. The evaluation metric is the standard mean Intersection-over-Union (mIoU). Specifically, we attach a linear classification layer to the patch tokens (after layer normalization) of the frozen encoder and train it on the ADE20k training set. We train with batch size as 16, learning rate as $1e-3$, weight decay as $1e-3$, $512\times512$ resolution, AdamW as the optimizer for 40k iterations.

\subsection{Dense Prediction: Depth Estimation}
We evaluate the performance of vision encoders in depth estimation using linear probing on the NYUv2 dataset \cite{nyu}. Results are reported using the Root Mean Squared Error (RMSE) metric (lower the better). We train a linear classifier on the training set. This linear layer is applied on top of the patch output features (after layer normalization) of the frozen encoder, with the features further normalized using a trained batch normalization layer. We train with batch size as 16, learning rate as $3e-4$, weight decay as $1e-3$, AdamW as the optimizer for 38k iterations.

\subsection{Dense Prediction: 3D Keypoint Matching}
We evaluate the performance of vision encoders in semantic correspondence on the SPair-71k dataset \cite{spair} in a training-free setting using a similar protocol as previous works \cite{walmer2023teaching, suri2024lift}. We use images resized to a side length of 448/512 pixels for models with patch size 14/16 respectively. 
Given an image pair with annotated source keypoints, we first extract dense feature maps from the frozen encoder for both images. For each source keypoint, its corresponding feature vector is obtained by bilinearly upsampling the feature maps to the image resolution and extracting the feature at the rounded keypoint pixel location. We then compute cosine similarity between this source feature and all spatial features in the target image to produce a similarity map, and predict the correspondence as the location with maximum similarity. The predicted location is mapped back to image coordinates and compared with the ground-truth target keypoint. Performance is measured using Percentage of Correct Keypoints (PCK), where a prediction is considered correct if the distance to the ground truth is within a specific pixel threshold of the object bounding box in the target image.
We choose 0.1 as the threshold (PCK@0.1), where the predicted point must be within 10\% of the maximum object bounding box dimension. We also set the number of image pairs per-category to 100 for our evaluation.

\clearpage
\newpage

\bibliographystyle{assets/plainnat}
\bibliography{main}

\end{document}